\pdfoutput=1

\documentclass[11pt]{article}

\usepackage[]{EMNLP2022}

\usepackage{times}
\usepackage{latexsym}

\usepackage[T1]{fontenc}

\usepackage[utf8]{inputenc}

\usepackage{amssymb}
\usepackage{bbm}
\usepackage{bm}
\usepackage{titlesec}
\usepackage{multirow}
\usepackage{hyperref}
\usepackage{mathtools}
\usepackage{colortbl}
\usepackage{float}
\usepackage{xcolor}
\usepackage{adjustbox}
\usepackage{enumitem}
\usepackage{arydshln}
\usepackage[capitalize]{cleveref}

\usepackage{booktabs}
\usepackage[font=small,labelfont=bf,tableposition=top]{caption}
\DeclareCaptionLabelFormat{andtable}{#1~#2  \&  \tablename~\thetable}

\usepackage{microtype}

\usepackage{inconsolata}

\title{MOCHA: A Multi-Task Training Approach for Coherent Text Generation from Cognitive Perspective}

\author{Zhe Hu \\
  Baidu Inc \\
  \texttt{huzhe01@baidu.com} \\\And
  Hou Pong Chan \\
  University of Macau \\
  \texttt{hpchan@um.edu.mo} \\\And
  Lifu Huang \\
  Virginia Tech \\
  \texttt{lifuh@vt.edu} \\
  }

\begin{document}
\maketitle

\begin{abstract}
Teaching neural models to generate narrative coherent texts is a critical problem. Recent pre-trained language models have achieved promising results, but there is still a gap between human written texts and machine-generated outputs. In this work, we propose a novel multi-task training strategy for coherent text generation grounded on the cognitive theory of writing, which empowers the model to learn essential subskills needed for writing including planning and reviewing besides end-to-end generation. 
We extensively evaluate our model on three open-ended generation tasks including story generation, news article writing and argument generation. Experiments show that our model achieves better results on both few-shot and fully-supervised settings than strong baselines, and human evaluations confirm that our model can generate more coherent outputs.

\end{abstract}

\section{Introduction}
With the recent development of pretraining techniques, 
large neural language models  have achieved impressive results on various text generation tasks and can generate fluent outputs. However, when generating \textit{long-form texts} (i.e., paragraphs with multiple sentences), there is still a large gap between machine-generated outputs and human written texts: the generated outputs usually suffer from incoherence issues and fail to maintain overall narrative coherence~\cite{see-etal-2019-massively}.

One possible reason for the above defects is the lack of effective text planning as global guidance to control the generation process. Compared with the traditional generation systems which often decompose the generation task into text planning and surface realization~\cite{reiter1997building,CARENINI2006925}, current autoregressive neural language models are typically trained to produce texts in a left-to-right token-level manner, which lacks anchored goal to constrain the generation process~\cite{fan-etal-2019-strategies}. Recent studies incorporate text planning into neural models by leveraging structured representations~\cite{goldfarb-tarrant-etal-2020-content,hua-wang-2020-pair} or latent variables~\cite{wang2022language,hu-etal-2022-planet} as high-level plans, but they need manually designed domain-specific plans or complicated supervision signals to train the model.

Another reason is the ineffective usage of the negative samples as contrasts to teach the model better distinguish between correct and incorrect targets. Negative samples are useful to enhance the model ability to generate better outputs~\cite{he-glass-2020-negative}. Recent work explores techinques such as contrastive learning~\cite{lee2020contrastive,su2022contrastive,an2022cont} and unlikelihood training~\cite{Welleck2020Neural,li-etal-2020-dont} to leverage negative samples for model training.

We draw our motivations from the \textbf{cognitive process theory of writing}~\cite{flower1981cognitive}: \textit{``Writing is best understood as a set of distinct thinking processes which writers orchestrate or organize during the act of composing''}. In particular, the basic mental process of writing includes \underline{planning}, \underline{translating} (surface realization), and \underline{reviewing}, where the reviewing process further involves \underline{evaluating} and \underline{revising} subskills. 
Current language models are typically trained to maximize the token-level loglikelihood and thus learn to acquire the writing skills all at once. However, as stated by 
\citet{bruce1978cognitive}, learning the whole set of task components for writing at once makes the learning process very hard, and they suggest that \textit{the intermediate tasks benefit to acquiring and exercising different writing subskills}.

\begin{figure*}[t]
    \centering
    \includegraphics[scale=0.62]{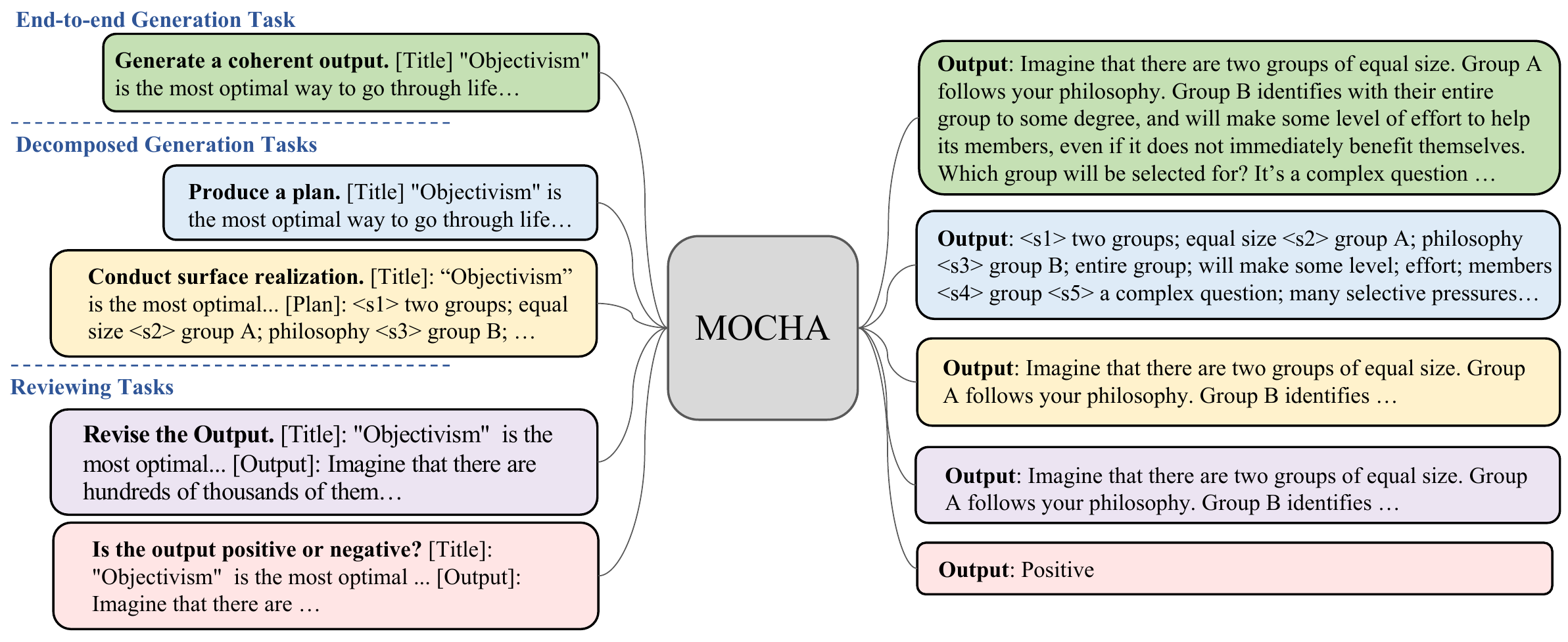}
    \vspace{-2mm}
    \captionof{figure}{ Overview of our framework. We train our model with different tasks grounded on the cognitive theory of writing: (1) end-to-end token-level generation task; (2) decomposed generation tasks including text planning and surface generation; (3) reviewing tasks with revising flawed targets and distinguishing between correct and incorrect options.
    }
    \vspace{-6mm}
    \label{fig:overall}
\end{figure*}

In this work, we propose \textbf{MOCHA}, a {M}ulti-task training appr{O}ach for {C}o{H}erent text gener{A}tion by enriching the token-level generation objective with additional tasks specifically designed for different writing subskills grounded on the cognitive perspective. Specifically, we introduce two additional tasks needed for generating coherent outputs: (1) decomposed generation tasks that divide the end-to-end generation into text planning and surface realization, and (2) reviewing tasks which leverage negative samples to enforce the model to distinguish the correct and incorrect outputs and further revise the flawed texts.

Our work is closely related to the recent multi-task training approach~\cite{sanh2021multitask} by converting different tasks into text-to-text transfer with corresponding prompts. Recent work~\cite{raffel2019exploring} has shown that multi-task learning (MTL) with shared parameters across different tasks can effectively improve the model performance on text understanding~\cite{aribandi2021ext5}, dialogue generation~\cite{li2021knowledge,su2021multi}, and structured knowledge grounding~\cite{xie2022unifiedskg}. 
Different from previous work, we study coherent long text generation with MTL to tackle different subskills needed for writing.
Experimental results show that our method outperforms strong baselines and achieves better few-shot performance compared with vanilla T5 on story generation, counter-argument generation and news article writing. Human evaluation further confirms that our method can generate more coherent outputs. Data and Code are available at: \url{https://github.com/Derekkk/Mocha-EMNLP22}

\section{Method}

Text generation is typically formulated as a sequence-to-sequence (seq2seq) transformation:
$p(y|x)=\prod_{t=1}^{n}p(y_t|y_{1:t-1}, x)$,
where $(x, y)$ is a source-target pair. We adopt the state-of-the-art model T5~\cite{raffel2019exploring} as the backbone, which is an encoder-decoder Transformer. For each sample, we introduce additional training objectives to jointly improve the writing ability. Our training objectives include  \textit{end-to-end generation}, \textit{decomposed generation} and \textit{reviewing} tasks. All tasks are converted to text-to-text transfer with a task prompt prepended to the source input. \textit{Notably, training samples of the augmented tasks can be constructed automatically, without further data labeling efforts}.
The overall framework is illustrated in Figure~\ref{fig:overall}.

\subsection{End-to-end Generation Task}
The end-to-end generation (Gen.) task is the same as the typical training objective for text generation. 
We prepend the source input with a task prompt (e.g., \textit{``Generate a coherent output''}), and the model is trained to generate the target. 
However, only applying this task is hard to generate coherent outputs as it couples the whole set of writing processes at once and makes training difficult. Therefore, we introduce the additional subtasks.

\subsection{Decomposed Generation Task}

Generating narrative coherent outputs requires the model to conduct effective \textit{text planning} to decide  high-level plots, and properly reflect the plans in the \textit{surface outputs}. Thus, we propose two decomposed generation tasks (Decomp.).

\smallskip
\noindent\textbf{Text Planning.} 
This task requires the model to produce \textit{structured plots} as high-level plans. We follow ~\citet{hua-wang-2020-pair} to adopt ordered keyphrase chains to represent the plots.
Concretely, we extract salient noun and verb phrases from the target as keyphrases, and then concatenate the keyphrases with the same order they appear in the target as the plan (more details are in Appendix~\ref{sec:plan_constrction}).
The task prompt \textit{``Produce a plan''} is prepended to the title, and the model is trained to generate the text plan, as shown in Figure~\ref{fig:overall}.

\smallskip
\noindent\textbf{Surface Realization.}
Surface realization task teaches the model to properly reflect the text plan in the final target. We concatenate the task prompt (e.g., \textit{``Conduct surface realization''}), title and the corresponding plan as the input sequence, which is consumed by the model to generate the final target.

\subsection{Reviewing Task}

We propose two reviewing (Review.) tasks  which leverage negative samples to enhance the model to better distinguish the coherent outputs from distracts, and learn to revise the flawed outputs.

\smallskip
\noindent\textbf{Revise Task.} The revise task aims to empower the model to edit the flawed outputs~\cite{wang-etal-2018-paper}. For each sample, we construct two flawed negatives: (1) randomly shuffle the target sentences to encourage model to learn correct sentence ordering, and (2) replace the keyphrases in the target with random keyphrases to enhance better content organization. The model takes as input the task prompt (\textit{``Revising the Output''}), title, and the flawed output, and recovers the original target.

\smallskip
\noindent\textbf{Distinguishing Task.}
This task requires the model to distinguish the original output from the distracted ones given an input. The distracted targets are constructed with the same strategies as the Revise Task. Similar to ~\citet{zhou2020pre}, the input sequence is the concatenation of the task prompt (e.g., \textit{``Which Option is Better''}), the title, an output with 50\% to be the original target or a distracted one otherwise. The model is trained to predict whether the output is correct by generating
\textit{``positive''} or \textit{``negative''}. 
By doing so, we expect the model to give a preference of the coherent targets and learn to generate better outputs.

\subsection{Joint Training with Multi-tasks}
We jointly train the aforementioned objectives with shared parameters to reinforce the writing ability. Specifically, given a source-target pair $(x, y)$, we first construct two decomposed generation samples for
text planning and surface realization tasks respectively. Then we construct two flawed samples for the revise task. 
Finally, for the distinguishing task, we choose the output with 50\% to be the positive target or a distracted negative target otherwise. 
All objectives are converted to text-to-text transfer tasks, and jointly trained to maximize the likelihood probability: $\mathcal{L} = \mathcal{L}_{\text{Gen.}} + \mathcal{L}_{\text{Decomp.}} + \mathcal{L}_{\text{Review.}}$. During inference, we use the end-to-end generation task to produce final outputs.

\section{Experimental Setting}
\begin{table}[t]
\fontsize{9}{12}\selectfont
 \setlength{\tabcolsep}{1.2mm}
  \centering
    \begin{tabular}{lccc}
        \toprule
        {} & {\bf Reddit/CMV} & {\bf Wikiplots} & {\bf NYTimes}\\
        \midrule
        \# Train & 42,462 & 95,571 & 103,579 \\
        \# Dev & 6,480 & 5,328 & 5,000 \\
        \# Test & 7,562  & 5,404 & 5,000 \\
        \# Words & 116.3 & 425.4  & 218.2\\
        \# Sent. & 5.5 & 18.0 & 9.1 \\
        \bottomrule
    \setlength{\abovecaptionskip}{0.5mm}
    \end{tabular}
    \vspace{-2mm}
    \caption{
    Statistics of the datasets. \# Words denotes the average number of words in the target, and \# Sent. represents the average number of sentences.
    }
    \label{tab:data_stat}
    \vspace{-6mm}
\end{table}

\subsection{Datasets} 
\label{section:data}
We evaluate our model on three datasets of distinct domains: (1) Reddit/ChangeMyView (Reddit/CMV) for argument generation~\cite{hua-wang-2020-pair}, (2) Wikiplots for story generation, and (3) New Tork Times for news article writing~\cite{sandhaus2008new}. We follow the previous work~\cite{rashkin-etal-2020-plotmachines} to further include topical keyphrases as guidance outline, where noun and verb phrases which contain at least one topic signature words~\cite{lin-hovy-2000-automated} from targets are extracted.
The title and keyphrases are concatenated as the input $x$. The statistics are in Table~\ref{tab:data_stat}, and 
more details are in Appendix~\ref{sec:data_details}.


\subsection{Model Details}
We use T5-base~\cite{raffel2019exploring} in all experiments. During training, we optimize our model with AdamW~\cite{loshchilov2017decoupled}, and the learning
rate is 5e-5. 
For decoding, we apply nucleus sampling~\cite{holtzman2019curious} with k as 10 and p as 0.9. The maximum of generation steps are 200 for argument generation, 512 for story generation and 350 for NYT article generation. 

\smallskip
\noindent\textbf{Baselines.}
We first consider generation models including GPT2~\cite{brown2020language} and T5~\cite{raffel2019exploring} without multitask training. We also include strong planning-based methods: 
(1) \textsc{ContentPlan} is a two-step generation model~\cite{goldfarb-tarrant-etal-2020-content,hua-wang-2020-pair}, where a planner first produces ordered keyphrase plans, and a generator consumes the plans and generates final outputs; (2) \textsc{BowPlan}~\cite{kang-hovy-2020-plan} predicts keywords as the global plan to guide the generation. All models are implemented with T5-base except for GPT2. More details are in Appendix~\ref{sec:exp_details_all}

\begin{table*}[t]
\fontsize{9}{12}\selectfont
 \setlength{\tabcolsep}{1.45mm}
  \centering
    \begin{tabular}{l cccc c cccc c cccc}
        \toprule
        & \multicolumn{4}{c}{\textbf{Reddit\//ChangeMyView}} & \phantom{} & \multicolumn{4}{c}{\textbf{Wikiplots}} 
        & \phantom{} & \multicolumn{4}{c}{\textbf{New York Times}}
        \\
        \cmidrule{2-5} \cmidrule{7-10} \cmidrule{12-15}
        \textbf{System} & \textbf{B-3} & \textbf{R-L} & \textbf{Meteor} &
        \textbf{Len.}  &
        \phantom{} &
        \textbf{B-3} & \textbf{R-L} & \textbf{Meteor} & 
        \textbf{Len.} &
        \phantom{} &
        \textbf{B-3} & \textbf{R-L} & \textbf{Meteor} & 
        \textbf{Len.} \\
        \midrule
        \textsc{GPT2} & 19.29  & 23.51 & 37.56  & 129 & \phantom{} & 11.39  & 20.00 & 26.91  & 299  & \phantom{} & 16.32 & 21.83 & 31.28 & 212 \\
        
        \textsc{BowPlan} & 27.19  & 26.86 & 44.33 & 109 & \phantom{} & 12.35  & 22.79 & 30.61 & 229 & \phantom{} & \underline{20.26}  & 25.40 & \underline{36.22} & 175 \\
        
        \textsc{ContentPlan} & 25.70  & 25.71 & 43.73 & 109 & \phantom{} & \bf{13.67} & 21.98 & \underline{32.16} & 260 & \phantom{} & 19.54  & 23.15 & 34.73 & 191 \\
        
        \textsc{T5} & 26.99   & 26.97 & 43.42 & 109 & \phantom{} & 11.99  & 23.08 & 30.27 & 221  & \phantom{} & 20.05  & 25.90 & 35.85  & 168 \\
        
        \hline
        \textbf{Our Models} \\
        \textsc{MOCHA} & \bf{28.02} & \bf{27.42} & \bf{44.81} & 110 
        & \phantom{} & {12.43} & \bf{23.43} & {30.94} & 224 &\phantom{}
        & \bf{20.43} & \underline{26.21} & \bf{36.45} & 166 \\
        
        \quad w/o Decomp. & \underline{27.60} & 27.28 & {44.41} & 108 & \phantom{} & 12.12 & 23.10 & 30.60 & 221 & \phantom{} & 19.80 & \bf{26.28} & 35.83 & 160 \\
        
        \quad w/o Review. & 26.92 & \underline{27.31} & 43.59 & 106 & \phantom{} & 11.36 & \underline{23.35} & 30.32 & 211 & \phantom{} & 19.97 & 26.20 & 36.07 & 164 \\
        
        \quad w/ SepGen. & 27.22 & 26.92 & \underline{44.67} & 103 & \phantom{} & \underline{13.54} &22.91 & \bf{32.26} & 249 & \phantom{} & 19.87 & 24.08 & 35.16 & 181 \\

        \bottomrule
    \end{tabular}
    \vspace{2mm}
    \caption{  
    Experimental results.
    We report BLEU-3, ROUGE-L (f), METEOR and average output lengths (Len.).
    Best results are bold and second best ones are marked with \underline{underline}.
  }
  \vspace{-8mm}
  \label{tab:auto-eval}
\end{table*}

\section{Results and Analysis}

\subsection{Automatic Results}
For automatic evaluation, we adopt BLEU~\cite{papineni-etal-2002-bleu}, ROUGE~\cite{lin-2004-rouge}, and METEOR~\cite{denkowski-lavie-2014-meteor}. 

The main results on three datasets are summarized in Table~\ref{tab:auto-eval}. Compared with the baseline methods, our model variants generate outputs with higher scores on all tasks. Specifically, our model significantly outperforms the vanilla T5 after augmenting the additional tasks, which demonstrates that our multitask training to tackle different writing subskills is critical to improve the model ability for long text generation. Moreover, our multitask training approach also surpasses or is comparable to the state-of-the-art planning-based methods, which further confirms the effectiveness of our method. We also observe that GPT2 achieves lower scores than other baselines, which indicates that only one decoder may not be sufficient since the tasks require the model to well understand and utilize the keyphrases during generation.

\subsection{Model Analysis}
\smallskip
\noindent\textbf{Ablation Study.} 
We train ablated model variants to analyze each augmented task. We first remove the decomposed generation tasks ({w/o Decomp.}), and the performance decreases, suggesting that incorporating text planning and surface realization tasks is helpful to improve long text generation. After removing the reviewing tasks (w/o Review.), the scores also drop, showing that improving model distinguishing and revise skills are useful to further enhance the generation ability\footnote{We further compute accuracy scores of the distinguishing task in Appendix~\ref{sec:eval_dist}.}.
Above all, the results prove that using different writing subskills to jointly train the model can effectively improve the overall end-to-end writing performance. 

\begin{figure}[t]
    \centering
    \includegraphics[scale=0.302]{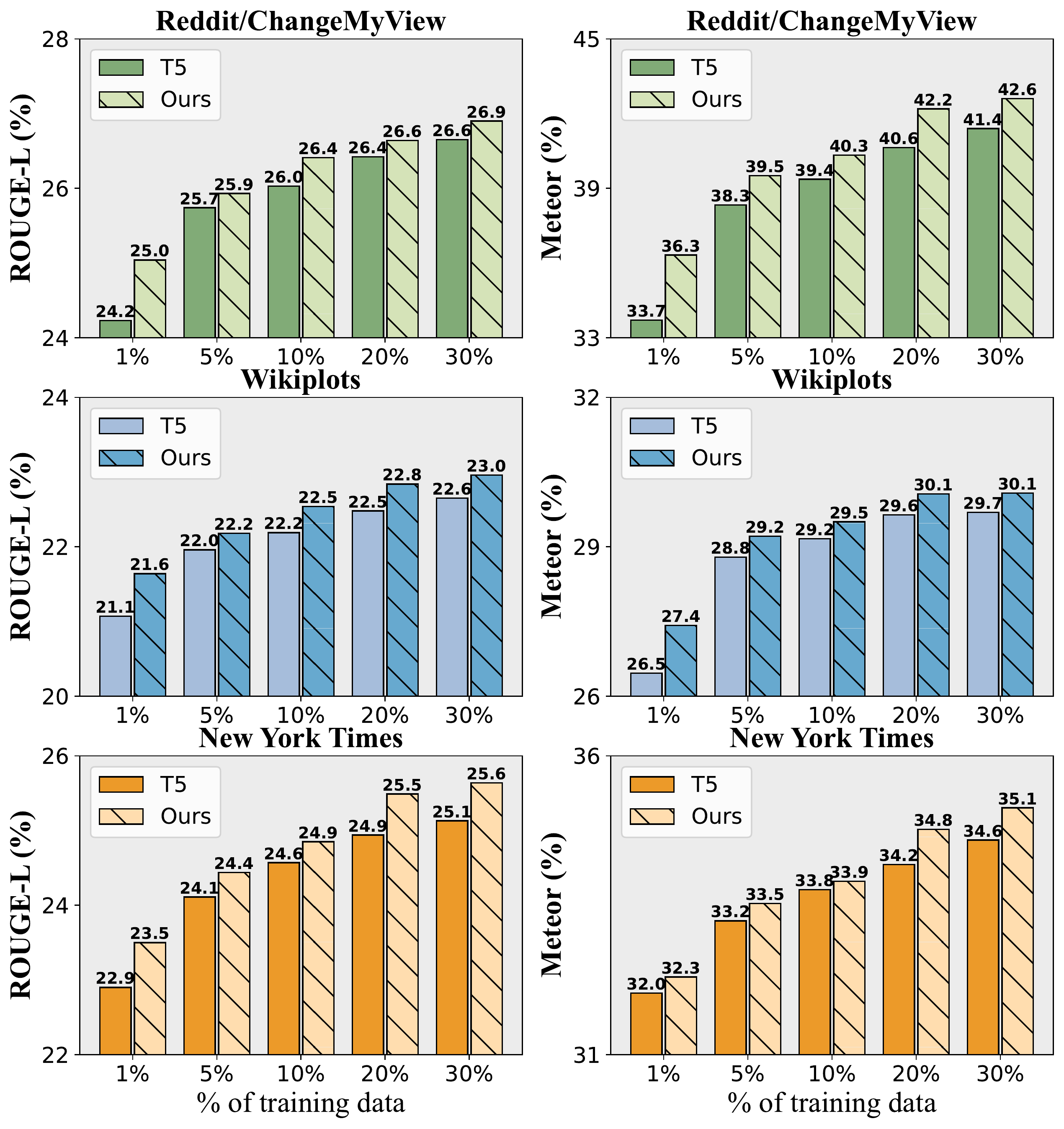}
    \vspace{-6mm}
    \captionof{figure}{ Few-shot results on test sets with different percentage of training samples. For each setting, we average the results of 5 runs with different sample selections.
    }
    \vspace{-5mm}
    \label{fig:few-shot-res}
\end{figure}

\begin{figure*}[t]
    \bgroup
    \def\arraystretch{1.8}
	\fontsize{10}{12}\selectfont
     \hspace{-1mm}
	\setlength{\tabcolsep}{0.8mm}
	\centering
	\begin{tabular}{p{155mm}}
	\toprule
        \vspace{-6mm}
	\textbf{Statement}: CMV: No one over the age of 80 should be allowed to serve in government.  \\
	\midrule	
        \vspace{-7mm}
	\textsc{\textbf{T5}}:  The problem is that the idea of a strict age limit for someone to serve is arbitrary. A competency exam is retesting on drivers licenses. It makes all sides feel like there is some kind of merit system. Having a 264-day old person in office is not good. The idea of making something so arbitrary that the best way to measure someone's competence is to have a competency exam. Derivising a competency exam in a political setting makes me happy. \\
	\hdashline
	\textsc{\textbf{MOCHA}}: The problem with the idea of a strict age limit is that it makes all sides look arbitrary. It's the idea that competency exams are a way to measure someone's competence, not some kind of merit system. It's like a 264-day old person devising a competency exam. I'm not sure I would be happy about that in a political setting. The idea that we need a competency exam to get to office makes me feel more competent. The idea of retesting on drivers licenses is pretty good, but that is not what's really the problem. 
	\\
    \bottomrule
	\end{tabular}
    \vspace{-2mm}
\caption{Sample outputs on counter-argument generation from Reddit/ChangeMyView test set. 
} 
\label{fig:samples1}
\egroup
\vspace{-3mm}
\end{figure*}

\smallskip
\noindent\textbf{Generation with Sub-tasks.} 
We also generate outputs with separate sub-tasks (w/ SepGen. in Table~\ref{tab:auto-eval}) rather than end-to-end generation during the inference time. Given an input, we first use text planning task to produce a plan, and then adopt the surface realization task and review task to generate and revise the final outputs~\footnote{We use the distinguishing task to decide whether revising the output or not.}, similar to \textsc{ContentPlan}. With MTL, our model variant significantly outperforms \textsc{ContentPlan}, which indicates the effectiveness of our MTL approach. Notably, separated generation leads to performance drops on Reddit and NYT compared to end-to-end generation. A possible reason is multi-step generation may bring cascading errors~\cite{castro-ferreira-etal-2019-neural}, and we leave this to future work.

\smallskip
\noindent\textbf{Few-shot Evaluation.}
We also conduct low-resource experiments to verify the effectiveness of our multitask learning approach. In particular, we vary the percentage of training data from 1\% to 30\%. To reduce variance, we repeat each experiment 5 times with different sample selections.
As shown in Figure~\ref{fig:few-shot-res}, compared with T5 without multitask training, our model consistently yields better results on all tasks. This verifies that our approach can learn general writing skills and be better adapted to new tasks with fewer samples.

\begin{table}[t]
\fontsize{9}{12}\selectfont
 \setlength{\tabcolsep}{1.5mm}
  \centering
    \begin{tabular}{lcc lc}
        \toprule 
        {\bf Task}  & {\bf Gram.} & {\bf Coh.} & {\bf Rich.} & {\bf Over.} \\
        \midrule
        Reddit/CMV  &  70.0\%  & 71.7\% & 66.7\% & 76.7\% \\
        NYT & 68.3\% & 73.4\% & 55.0\% & 65.0\% \\
        Wikiplots & 86.7\% & 86.7\% & 63.3\% & 86.7\% \\
        \bottomrule
    \end{tabular}
    \vspace{2mm}
    \caption{
    Human evaluations. We report (averaged) \% of times our model are considered better than T5 on grammaticality (Gram.), coherence (Coh.), content richness (Rich.) and overall quality (Over.). All Krippendorff’s $\alpha \geq 0.31$.  
  }
  \label{tab:human_eval_res}
  \vspace{-7mm}
\end{table}

\subsection{Human Evaluation}
We hire two proficient English speakers as human annotators to evaluate model outputs. We randomly choose 30 sample inputs from test set per task. For each input, we anonymously present the outputs of T5 and our model, and ask each annotator to choose the better output based on: (1) grammaticality; (2) coherence; (3) content richness and (4) overall quality. The final results are shown in Table~\ref{tab:human_eval_res}. As we can see, human judges consider our model outputs better than T5 on all aspects. In particular, over 70\% times our model results are regarded more coherent, which confirms that our multitask learning approach can effectively improve the output coherence. Moreover, our multitask training is more effective on Wikiplots where the outputs are longer, indicating its effectiveness for long-form story generation.
Among all aspects, the smallest gap is observed on content richness, which suggests the future directions of designing more specific tasks to improve output diversity.

In Figure~\ref{fig:samples1}, we present a sample output on argument generation from Reddit dataset. 
Compared with vanilla T5 without multi-task training, our model is able to produce more coherent and relevant outputs with correct stance (a counter-argument aims to refute the statement). In contrast, the sentences in
T5 output are less cohesive, and some sentences suffer from incorrect stance (e.g., \textit{``Having a 264-day old
person in office is not good''}). Additional sample outputs on other tasks are presented in Appendix~\ref{sec:sample_output}.

\section{Conclusion}
In this work, we present a multitask training approach driven by cognitive theory of writing to improve the model ability on coherent long text generation. We introduce decomposed generations tasks and reviewing tasks with different prompts to tackle essential subskills needed for generating coherent outputs. Our model achieves better results on three long text generation tasks under both full training and few-shot settings, and can generate better and more coherent outputs.

\section*{Acknowledgements}
We thank the anonymous reviewers for their valuable
suggestions. Hou Pong Chan was supported by the Science and Technology Development Fund, Macau SAR (Grant No. 0101/2019/A2), and the Multi-year Research Grant from the University of Macau (Grant No. MYRG2020-00054-FST). 

\section*{Limitations}
Training neural language models to generate coherent long-form outputs is an important task. In this work, we improve model writing ability with multitask training based on the cognitive theory. Nevertheless, we believe there is still huge space to explore in the future. First, in this work we apply our multitask training on each downstream tasks, while one could apply our approach in the pretraining stage to obtain a general writing model. Our proposed decomposed generation and reviewing tasks do not require additional labeled data, and the data can be constructed automatically. Thus applying our approach in the pretraining stage could be useful to improve the model ability. Second, in this work we adopt ordered keyphrases as planning plot. However, for different writing tasks, there might be different ways to represent the plot such as using semantic role labels~\cite{fan-etal-2019-strategies} or entity chains~\cite{narayan-etal-2021-planning}. Future work might incorporate different plot representations on downstream tasks to further boost the model performance.

\section*{Ethics Statement}
In this work, we study long text generation task.
We recognize that our method may generate contents which contain potentially harmful information and malicious languages due to the systematic biases of pre-training with web corpora. Therefore, we urge the users to carefully check the model outputs, examine the ethical influence of the generated contents, and cautiously deploy the system in real applications. 

\bibliography{anthology,custom}
\bibliographystyle{acl_natbib}

\appendix

\section{Experimental Details}
\label{sec:exp_details_all}

\subsection{Datasets}
\label{sec:data_details} 
For data preprocessing, we keep the original texts and do not lowercase the tokens. For topical keyphrases, we extract noun and verb phrases which contain at least one topic signature words~\cite{lin-hovy-2000-automated} from the targets.

\smallskip
\noindent\textbf{Argument Generation.} 
The first task requires the model to generate a counter-argument given a statement on a controversial topic. We adopt the Reddit/ChangeMyView dataset processed by ~\citet{hua-wang-2020-pair}. The data are collected from Reddit ChangeMyView subcommunity, where the original post title are considered as the input, and the replies are considered as the target counter-arguments. The noun and verb phrases which contain at least one topic signature words~\cite{lin-hovy-2000-automated} are extracted from the targets to serve as the topical keyphrases~\cite{hua-etal-2019-argument-generation}.

\smallskip
\noindent\textbf{Story Generation.}
For story generation, we apply Wikiplots dataset~\footnote{https://github.com/markriedl/WikiPlots}, which consists of story plots of disfferent genres such as TV shows,
movies, and books, scraped from Wikipedia. We use the processed dataset from ~\citet{ji-huang-2021-discodvt}. We use the same way as in Argument Generation to extract the outline keyphrases.

\smallskip
\noindent\textbf{News Article Writing.}
For news article writing, we consider the articles from New York Times dataset~\cite{sandhaus2008new}. We apply the processed data by ~\citet{hua-etal-2021-dyploc}. In their original dataset, entities and concepts  extracted from external knowledge base are considered as additional inputs. In our setup, we ignore the knowledge items and instead extract keyphrases as outlines the same as in Argument Generation. The data splits are the same as the original paper.

\subsection{Text Plan Construction}
\label{sec:plan_constrction}
For the planning task, we represent the output plot with ordered keyphrase chain. Specifically, we consider all input topical keyphrases, as described in Section~\ref{section:data}. Then we concatenate all keyphrases with the same order they appear in the target as text plan.
For the $i$-th story sentence, the keyphrase chain is: ``$k_{i1};k_{i2};...;k_{im}$'', where $k_{im}$ is the $m$-th keyphrase appeared in the $i$-th sentence. We then concatenate keyphrase chains of all sentences with a special token <sep> as the ordered keyphrase chain. For example: ``$k_{11};k_{12};...\text{<sep>}k_{21};k_{22}...\text{<sep>}...$''.


\subsection{Training Details}
\label{sec:exp_details}
We use T5-based~\cite{raffel2019exploring} in all experiments. During training, we set the maximum length of both input and output as 512. We implement all experiments using the Huggingface Transformers~\cite{wolf2020transformers} and Pytorch~\cite{paszke2019pytorch}. The maximum training epoch is set as 12 for argument generation and 18 for story generation and article writing. We optimize our model with AdamW~\cite{loshchilov2017decoupled}. The batch size is 8, and the learning rate is 5e-5. For our multi-task training, we first construct augmented samples for all sub-tasks using the method as described previously. The original training instances and the augmented samples are then mixed together as the new training set to train the model.

For decoding, we apply nucleus sampling~\cite{holtzman2019curious} with k as 10 and p as 0.9. The maximum of generation steps are 200 for argument generation, 512 for story generation and 350 for NYT article generation. We use NVIDIA V100 GPUs for all experiments, and the best model checkpoint is chosen based on the validation loss. It takes roughly 6 hours to converge for argument generation, 20 hours for article generation, and 24 hours for story generation with 8 GPUs. Our model size is the same as vanilla T5 base version.

\subsection{Baselines and Comparisons}
For comparison, we first consider strong generation models including GPT2~\cite{brown2020language} and T5~\cite{raffel2019exploring} without multitask training. We also include planning-based methods: 
(1) \textsc{ContentPlan} is a two-step generation model~\cite{goldfarb-tarrant-etal-2020-content,hua-wang-2020-pair}, where a planner first produces ordered keyphrase plans, and a generator consumes the plans and generates final outputs; (2) \textsc{BowPlan}~\cite{kang-hovy-2020-plan} predicts keywords as a global plan to guide the generation. All models are implemented with T5-base except for GPT2. 

\smallskip
\noindent\textbf{\textsc{BowPlan.}}
We construct the bag-of-words (BOW) label with content words of the target. Following ~\citet{kang-hovy-2020-plan}, the BOW planning distribution is incorporated at each decoder time step with a gated probability to compute the final output. 

\smallskip
\noindent\textbf{\textsc{ContentPlan.}}
This is a two-step generation method, where a planning model first produce the ordered keyphrase plans given an input, and then a generation model produces the final surface form output given the title and content plans. Both the planner and generator are initialized with T5-base. During training, we use gold plans to train the generator, and use the predicted plans during inference.
For decoding method, we apply nucleus sampling for both the planner and the generator.

\begin{table}[t]
\fontsize{10}{12}\selectfont
 \setlength{\tabcolsep}{1.9mm}
  \centering
    \begin{tabular}{lcc lc}
        \toprule 
        {\bf Task}  & {\bf Accuracy}  \\
        \midrule
        Reddit/ChangeMyView  &  83.4\% \\
        NYT & 98.3\% \\
        Wikiplots & 94.6\% \\
        \bottomrule
    \end{tabular}
    \vspace{2mm}
    \caption{
    Accuracy of the distinguishing task on each dataset.  
  }
  \label{tab:distingush_results}
  \vspace{-7mm}
\end{table}

\subsection{Evaluation on Distinguishing Task}
\label{sec:eval_dist}
We compute accuracy of  the Distinguishing Task on each dev set to analyze whether our model essentially learns the corresponding sub-skill. The results are shown in Table~\ref{tab:distingush_results}. All accuracy scores are above 80\%, which proves that our model is able to accurately distinguish between positive and negative targets. Notability, the result on Reddit is lower than results on NYT and Wikiplots, which is consistent with the conclusion from previous work~\cite{hu-etal-2022-planet}, as Reddit data are collected from social network and usually contain more informal expressions and noises.

\subsection{Details for Human Evaluation}
For human evaluation, we hire two proficient English speakers as human annotators. For each task, we randomly select 30 samples from the test set, and present the outputs of T5 and our MTL method, with the model names anonymized to reduce bias. We ask each human annotator to select the better outputs based on the following aspects: (1) \textbf{Grammaticality} to measure correct language usage; (2) \textbf{Coherence}, measuring whether the text has proper high-level plot and is cohesive; (3) \textbf{Content Richness} to measure the diversity and informativeness of outputs; and (4) \textbf{Overall Quality}.

\section{Sample Outputs}
\label{sec:sample_output}
We present additional sample outputs on news article writing and story generation from Figure~\ref{fig:samples2} to Figure~\ref{fig:samples3}.

\begin{figure*}[t]
    \bgroup
    \def\arraystretch{1.8}
	\fontsize{10}{12}\selectfont
     \hspace{-1mm}
	\setlength{\tabcolsep}{0.8mm}
	\centering
	\begin{tabular}{p{150mm}}
	\toprule
	\textbf{Title}: Unhappy Meals \\
	\midrule
	
	\textsc{\textbf{Human}}: Michael Pollan contends that lawsuits filed by the obese against fast-food chains"seem absurd on their face," because"no one's forcing people to eat this stuff." Literally, this is true. But more and more often these days, when we are obliged to eat away from home, the only alternative to fast food is hunger. As Critser reports in his book, the nutritious public school cafeteria lunch has become largely budget-cut out of existence, and the fast-food chains have taken up the slack. Highway food is now virtually synonymous with junk food. Airport lounge eateries dispense enough grease to lubricate the Concorde. And let's not even mention malls, where you could purchase the Maltese Falcon more easily than you could buy an apple not engulfed by caramel. When I was a kid in the early 1960's, our town's only McDonald's prominently displayed a big sign indicating how many hamburgers had been served to date, nationwide, under the Golden Arches. When"8 million" was at long last updated to"10 million," we were awed. My inexpert estimate would be that this figure is nowadays surpassed in a single weekend. \\
	
	\hdashline
	
	\textsc{\textbf{T5}}:  It was the nutritious public school cafeteria lunch that got me so excited I sat on the subway for the second straight night, and I was engulfed by caramel on a hot summer day that I could purchase the Maltese Falcon. But I was astonished by how many hamburgers I could eat at airport lounge eateries. I had the slack for my lunch at McDonald's and the McDonald's sandwich. And I'm sure I have surpassed in a single weekend the number of complaints filed by the obese and the overweight. But I was surprised at how many of the fast-food chains are not dispense enough grease to lubricate the Concorde and others, not to mention malls and airport lounge eateries. I've been eating this stuff for years, and it has never been synonymous with highway food. \\

	\hdashline
	\textsc{\textbf{MOCHA}}: I am awed by the number of cases of junk food filed by the obese, and awed that the nutritious public school cafeteria lunch is being surpassed in a single weekend. I can't get enough of it. The same could be said of the fast-food chains in New York City: airport lounge eateries engulfed by caramel and the slack of the Golden Arches, which lubricate the Concorde and eat this stuff for lunch. Not to mention malls and airports that dispense enough grease, but the fact is that many hamburgers aren't junk food.
	\\
    \bottomrule
	\end{tabular}
    \vspace{-3mm}
\caption{Sample outputs on NYT.
} 
\label{fig:samples2}
\egroup
\vspace{-4mm}
\end{figure*}

\begin{figure*}[t]
    \bgroup
    \def\arraystretch{1.8}
	\fontsize{10}{12}\selectfont
     \hspace{-1mm}
	\setlength{\tabcolsep}{0.8mm}
	\centering
	\begin{tabular}{p{150mm}}
	\toprule
	\textbf{Title}: Ultimates \\
	\midrule
	
	\textsc{\textbf{Human}}: General Nick Fury ofHIELD. establishes a strike force of government-sponsored metahumans which includes Captain America; scientist couple Henry and Janet Pym (Giant-Man and the Wasp); Bruce Banner (the Hulk) and Tony Stark (Iron Man). Together they are based at theHIELD facility, the Triskelion. When Banner injects himself with the super-soldier serum and goes on a bloody rampage as the Hulk, he is eventually stopped by the other metahumans with the aid of Thor. The team then join forces with the mutants Quicksilver and Scarlet Witch and agents Hawkeye and Black Widow against the alien shape-shifters the Chitauri, who are defeated. A year later public opinion has turned against the team when it is discovered that Bruce Banner is in fact the Hulk and was responsible for hundreds of deaths. The team is undermined further when Thor is accused of being an escaped mental patient and is incarcerated. This is the doing of his brother Loki, who also facilitates the creation of a new team of anti-American multi-nationals called the Liberators. With the aid of the Black Widow – who betrays the team to the Liberators – the Ultimates are captured, but eventually escape and battle the Liberators to the death (...) \\
	
	\hdashline
	
	\textsc{\textbf{T5}}:  Following the events of the first game, the Ultimates defeat a new team of asgardian warriors called the Liberators. The team is led by the `Booty-Horse', the Black Widow of the Universe, and the mutants Quicksilver, The Hulk and the Black Widow. The Liberators are led by a scout called the `Booty-Horse' and the Black Widow. The Ultimates also battle the Liberators, who are based at the HIELD facility. The Ultimates' leader, Tony Stark, betrays the team's leader, Tony Stark, who was turned against the team, with the aid of the Scarlet Witch, the Wasp and the Triskelion. Tony Stark is unable to distract from the robot's presence and the wasp becomes the target of the liberators, but Tony manages to defeat him. Tony and Tony's brother Loki are killed when the Triskelion switches his mind with a weapon that facilitates the creation of a new race called the Liberators. The Avengers are defeated by a gang of Asgardian warriors led by the mysterious dr Griswold, who uses the Ultimates to destroy the mutants. Tony is then attacked by a band of Ultron robots drugs and kills him. Tony is later stopped by the other Metahumans. Tony and Tony are then taken to theUniversals Mansion. Tony, Tony, and the Black Widow are reunited, with Tony's help, but Tony is unable to bring them back, as Tony is now being beaten up. Tony, Tony, and Tony are taken to theUniversals. \\

	\hdashline
	\textsc{\textbf{MOCHA}}: The Ultimates defeat the forces of the Scarlet Witch and the Triskelion, and the Ultimates are defeated in a final battle with the Wolverine. In the meantime, a new team called the Liberators, based at the HIELD facility, have taken the throne. However, a series of events are going on in the Ultimates mansion, and the Ultimates are stopped by the other metahumans. During the battle, the Ultimates are attacked by the `Red Hat', a race of government-sponsored metahumans, who were turned against the team due to the team's lack of training, and the Ultimates are defeated by the `Black Widow'. In an effort to stop the 'Red Hat', General Nick Fury of HIELD. agents Hawkeye and Hank Pym are unable to stop it and are ordered to leave. During their stay, Ultron robots drugs the team, and they retreat to a cave. Upon waking up, they discover that `the wasp' had turned against the team and they are unable to get back to the original one. The team then encounters the `Skunks', the mutants Quicksilver and Iron Man, and their mutants The Scarlet Witch and the Wasp. Meanwhile, at the headquarters of `The Black Widow', Janet Pym, who is the wife of Tony Stark, betrays the team by telling them that her brother Loki is in a coma. The team then goes to `The Black Widow's' house, where they fight the Liberators. The Liberators have discovered that Bruce Banner is actually an agent of the United States government, and have been trying to find a way to take over HIELD. headquarters. Tony Stark, with the aid of Asgardian warriors and shifters the Chitauri, facilitates the creation of the Triskelion. The Liberators are then defeated by the Black Widow, and Tony's `Unit' team is defeated.
	\\
    \bottomrule
	\end{tabular}
    \vspace{-3mm}
\caption{Sample outputs on Wikiplots.
} 
\label{fig:samples3}
\egroup
\vspace{-4mm}
\end{figure*}

\end{document}